\documentclass[10pt,twocolumn,letterpaper]{article}

\usepackage{algorithmicx}
\usepackage{algorithm}
\usepackage{booktabs}
\usepackage{cvpr}
\usepackage{graphicx}
\usepackage{algpseudocode}
\usepackage{booktabs}

\renewcommand{\algorithmicrequire}{\textbf{Input:}}
\renewcommand{\algorithmicensure}{\textbf{Output:}}

\cvprfinalcopy % *** Uncomment this line for the final submission

\begin{document}

%%%%%%%%% TITLE
\title{Bag Reference Vector for Multi-instance Learning$^\star$}

\author{Hanqiang Song$^{\ast\dagger}$, Zhuotun Zhu$^\ddagger$ and Xinggang Wang$^\ast$\\
$^\ast$Huazhong University of Science and Technology, \\$^\dagger$University of Birmingham, $^\ddagger$University of California, Los Angeles \\
{\tt\small hanqiangsong1024@gmail.com, zhuotun@gmail.com, xgwang@hust.edu.cn}
}

\maketitle
%\thispagestyle{empty}

%%%%%%%%% ABSTRACT
\begin{abstract}
Multi-instance learning (MIL) has a wide range of applications due to its distinctive characteristics. Although many state-of-the-art algorithms have achieved decent performances, a plurality of existing methods solve the problem only in instance level rather than excavating relations among bags. In this paper, we propose an efficient algorithm to describe each bag by a corresponding feature vector via comparing it with other bags. In other words, the crucial information of a bag is extracted from the similarity between that bag and other reference bags. In addition, we apply extensions of Hausdorff distance to representing the similarity, to a certain extent, overcoming the key challenge of MIL problem, the ambiguity of instances' labels in positive bags. Experimental results on benchmarks and text categorization tasks show that the proposed method outperforms the previous state-of-the-art by a large margin.
\end{abstract}

\footnotetext[1]{This work is under the consideration of Patten Recognition Letter.}

%\linenumbers

%% main text
\section{Introduction} \label{section:Intro}

Multi-instance learning (MIL), originally proposed for drug activity prediction \cite{dietterich1997solving}, has been applied more frequently to diverse visual recognition tasks such as image retrieval, image classification, object detection, and visual tracking. In MIL, a typical weakly-supervised learning, training data are given as a form of labeled bags, each of which is composed of a wide diversity of instances associated with input features. The aim of MIL, in a binary task, is to train a classifier to predict the labels of testing bags, which is based on the assumption that a positive bag contains at least one positive instance while a bag is labeled negative if it is only constituted of negative instances. Thus, the crux of MIL is to deal with the ambiguity of instances' labels, especially in positive bags which have plenty of cases with different compositions.

In a way, this weakly labeled instance framework cater to many existing visional tasks such as object recognition task for the reason that intrinsic structure of MIL is able to deal with some problems perfectly hence facilitate solutions. Take image classification for instance, an image is defined as a bag and patches in the image can  be regarded as instances. Then according to the purpose of MIL, specific objects or key features can be defined as positive. By means of this MIL representation, crucial information can be captured.

Up to now, different algorithms have been designed to solve MIL problems. The previous methodologies are mainly in three folds: (1) Selecting key/discriminative instances and classifying bags based on the selected instances using generative or discriminative models, e.g., EM-DD~\cite{Zhang01em-dd:an}, miSVM~\cite{andrews2002support} and the key instance detection method \cite{liu2012key}. (2) Mapping a bag into a high-dimensional feature space to get a vector representation of bag then training bag classifier, e.g. miFV~\cite{wei2014scalable}. (3) Constructing bag representation based on the internal structure of a bag - the relation between instances within a same bag, e.g., miGraph~\cite{zhou2009multi}.

Different from the previous strategies, we aim to build bag-level representation based on the relative distance between bags.
Take one bag for instance, other bags are regarded as reference bags functioned as the basis of the feature space.
The derived bag representation is called bag reference vector (BRV). Then MIL task is transformed into a problem of classifying BRV. Thus, our method to solve this task  is named miBRV (multi-instance learning of bag reference vector) as a whole.

Our motivation of proposing BRV is that it can capture the essential character of MIL.
For supervised learning, the intra-class similarly should be higher than the inter-class similarity.
In the same manner, the similarities between bags are distinctive features.
To measure bag similarity, we use set-to-set distance which considers all pairwise relations between two bags.
We consider all pairwise distances between the instances in the compared bags, which is a typical set-to-set distance.
In this paper, we extent Hausdorff distance \cite{huttenlocher1993comparing} as the set-to-set distance measure.
Furthermore, considering the ambiguity of instances in bags, we adopt a range of operators in Hausdorff distance to represent these relations.

\begin{figure*}
\centering
\includegraphics[width=0.7\textwidth]{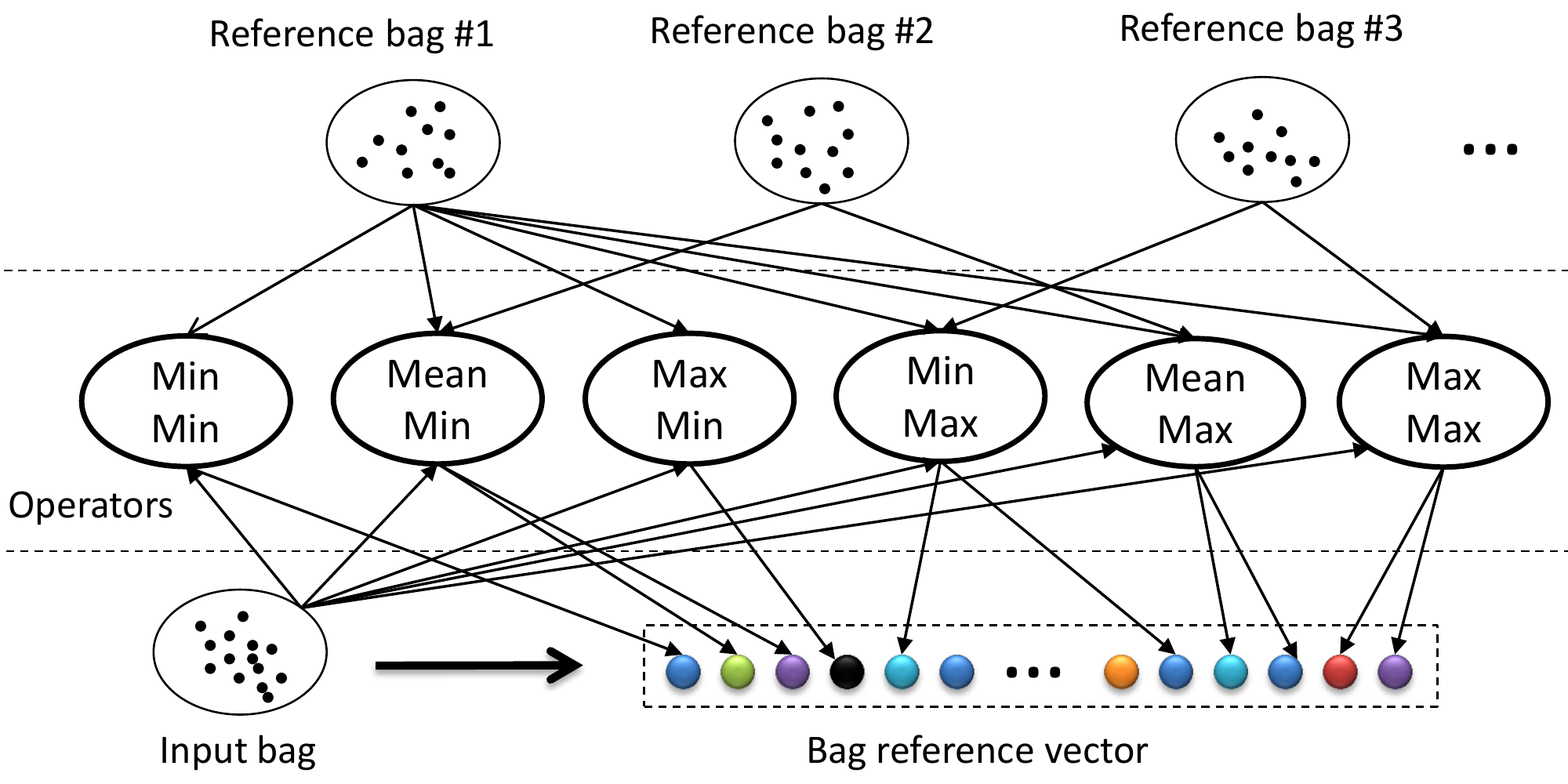}
\caption{The illustration of how to build bag representation. The first row shows some reference bags; the second row shows some operators which take two bags and compute a distance;
and the third row shows an input bag and its corresponding bag reference vector. Each dimension of the feature vector corresponds to a triplet \{input bag, an operator, reference bag \#n\}.}
\label{figure:Pipeline}
\end{figure*}

The pipeline of generating bag feature is illustrated in Fig.~\ref{figure:Pipeline}. For every reference bag and operator, the input bag has a
distance value to the reference bag based on the operator measurement. Total length of the bag reference operator is the product of
the number of reference bag and the number of operators.

In the rest of this paper, we briefly review some related works in Section~\ref{section:rw};
then formalize the proposed miBRV method for MIL in Section~\ref{section:Method};
in Section~\ref{section:Experiments}, we carry out experiments of miBRV on MIL benchmarks and show the state-of-the-art performance;
finally, we draw conclusions in Section~\ref{section:Con}.

\section{Related Work}  \label{section:rw}

Multi-instance learning (MIL) has  received a lot of attentions since it helps to solve a range of real applications. Till now, lots of MIL methods have been proposed to either develop effective MIL solvers or apply MIL to solve application problems.
Firstly, we briefly review a few popular MIL solvers.
The EM-DD method \cite{Zhang01em-dd:an} uses EM to infer instance space with many instances from different positive bags and few instances from negative bags.
Instead of adopting simple instance space, the miRPCA method \cite{wang2013one} utilizes robust PCA model to build a instance model robust to outliers.
Besides of generative instance models, discriminative models are more popular as instance model.
For example, both MILBoost \cite{Viola06multipleinstance} and miSVM use discriminative methods, Boosting and SVM respectively, as instance models, and iteratively select positive instances to train models.
Furthermore, miGraph \cite{zhou2009multi} represents bag as graph and explicitly model the relationships between instances within a bag;
while \cite{Deselaers_aconditional} models the relationships between different bags using conditional random field.
Recent work \cite{Babenko} studies the problem if there are infinite number of instances in a bag.

MIL is useful for many computer vision applications. Originally, MIL is widely applied to image classification \cite{maron1998multiple, zha2008joint}, since it is able to exploit salient region in image where is critical for classification. In \cite{felzenszwalb2010object}, a variant of miSVM called latent SVM is effective to find the parts of object for accurate object detection.
Online MIL algorithms \cite{babenko2011robust,zhang2013real} are popular for visual tracking. Recently, MIL has been widely used for weakly-supervised object detection \cite{wang2014robust, cinbis2014multi}.

In addition, our miBRV method is a reference-based method, which is analog to the popular concept ``Attribute" \cite{parikh2011relative} in computer vision.
A face feature computed based on reference face is proposed in \cite{shen2013face}. However, \cite{shen2013face} is simpler and there is no MIL structure in it.

\section{Multi-instance Bag Reference Vector} \label{section:Method}

In this section we will illustrate our bag-reference-based method applied to MIL problem. The miBRV aims to construct a vector
representation for each bag by computing the similarity (distance in our method) with all other bags which are taken as the reference,
transferring original features (with complex structure) into new bag-reference features containing rich information with simple linear structure.
Our intuition is to use the distances with the reference bags to describe the bags' intrinsic constitution and then train this affinity matrices to gain map function by a linear SVM.

\subsection{Multi-instance Learning}

Initially, we introduce the formal formulations of Multi-instance Learning. Given a data set of $\left\{(X_1,Y_1),...,(X_i,Y_i),...,(X_N,Y_N)\right\}$, where $N$ is the number of bags, each bag $X_i$ is consisted of grouped instances $\left(x_{i1},..., x_{ij},...,x_{in}\right)$ and labeled with $Y_i\in\{1,-1\}$ while the instances' labels are unknown.
A positive bag contains at least one positive instance while there are only negative instances in negative bags. Thus, the task of Multi-instance Learning is to induce a classifier (or a mapping function) to predict the labels of input bags.

\subsection{Bag Reference Vector}

As what have been mentioned in section \ref{section:Intro}, intending to measure the relations between bags by means of set-to-set distances, we apply an operator to represent these distances. Hausdorff distance is a suitable technique to determine the extent to which one bag differs from another.

\subsubsection{The Hausdorff Distance}

Given two point sets A = \{$a_1,...,a_m$\}, B = \{$b_1,...,b_n$\}, the Hausdorff distance is defined as

$$H(A,B) = max(h(A,B),h(B,A)),$$
where
$$h(A,B)= \mathop{max}_{a\in A} \mathop{min}_{b\in B}||a_i-b_i||.$$
Here the function $h(A,B)$, directed Hausdorff distance, is called forward Hausdorff distance from set $A$ to set $B$ as well.
In addition, $||a_i-b_i||$ represents Euclidean distance between $a_i$ and $b_i$, i.e., a point-to-point distance $d_{poi-poi}(a_i, b_i)$.
For each $a_i\in A$, the algorithm will compute the point-point distance from $b_1$ to $b_n$ and find the nearest point in set $B$ to $a_i$ with the least Euclidean distance,
which is regarded as a point-set distance $d_{poi-set}(a_i, B)$.
Hence, from set $A$ to $B$, we are able to gain a distance vector $[ d_{poi-set}(a_1, B)$,...,$d_{poi-set}(a_m, B) ]$
with $m$ dimensions and vice versa as it is asymmetric distance.
According to the definition for function $h$, the biggest one among these $m$ shortest distances will be select as the value of $h(A,B)$,
representing the distance form $A$ to $B$.
In this way we define a method to compute the set-to-set distance, in other words, this measures the similarity between set $A$ and set $B$.

In terms of MIL, similar definition can be applied to it. Naturally, we treat each bag as a set like $A$ or $B$ and instances as the points. Thus, for bag $X_i$ and $X_j$, we can apply forward Hausdorff distance to it as $$h(X_i,X_j)= \mathop{max}_{x_i\in X_i} \mathop{min}_{x_j\in X_j}||x_i-x_j||$$ For bag $X_i$, we can obtain a bag-reference vector
$$b^{X_i} = \left[ h(X_i, X_1 ),...,h(X_i, X_j ),...,h(X_i, X_N) \right]$$
where $j=1, 2, ..., N$.
In addition, $b^{X_i}$ is $l_2$  normalized to reduce the influence of instance magnitude variation.
With each bag's vector computed, an affinity matrices $b^X$ can be extracted by means of Hausdorff distance. Then each bag $b^X$is delineated by being compared with the reference bags.
The feature matrices are fed into a bag classifier along with bag labels for training and validation.

Pseudo code of miBRV is shown in Algorithm~\ref{alg:1}.

\subsubsection{Extensions of Hausdorff Distance}
The Hausdorff distance defines a point-to-set distance $d_{poi-set}$ by finding the nearest point in that set with the least Euclidean distance and then chooses the maximum among these point-to-set distances as final set-to-set distance. Thus, the operator is to obtain the maximal one among minimum values. This operation is suitable to most cases to gain correct descriptions for positive and negative bags.

 Furthermore, considering the characteristic of MIL that some positive bags may well include negative instances, there will be some flaws in this algorithm  for all multiple cases.
For instance, if $X_i$ is a positive bag with one negative instance $x_{i1}$ while $X_j$ is composed of positive instance only, after using Hausdorff distance, the bag-to-bag distance will be equal to the maximal instance-to-bag distance in
$[ d_{ins-bag}(x_{i1}, X_j)$,...,$d_{ins-bag}(x_{im}, X_j) ]$, which is $d_{ins-bag}(x_{i1}, X_j)$ as $x_{i1}$ has the largest distance with all instances in $X_j$ . This indicates using the Hausdorff distance between two positive bags results in choosing the distance from negative instance to positive bags to represent the similarity, which gives the misleading information.
Consequently, modifications can be adopted in Hausdorff distance to gain several new affinity matrices as complements.
Specifically, maximum, average and minimum operators have been added to enrich the distance definition to ameliorate incorrect representation.
The following illustrates Hausdorff distance as well as other five distance measurement operators paralleled to it.

$$h_{minmin}^1(X_i,X_j)= \mathop{min}_{x_{im}\in X_i} \mathop{min}_{x_{jn}\in X_j}||x_{im}-x_{jn}||.$$
$$h_{meanmin}^2(X_i,X_j)= \mathop{mean}_{x_{im}\in X_i} \mathop{min}_{x_{jn}\in X_j}||x_{im}-x_{jn}||.$$
$$h_{maxmin}^3(X_i,X_j)=h(X_i,X_j)= \mathop{max}_{x_{im}\in X_i} \mathop{min}_{x_{jn}\in X_j}||x_{im}-x_{jn}||.$$
$$h_{minmax}^4(X_i,X_j)= \mathop{min}_{x_{im}\in X_i} \mathop{max}_{x_{jn}\in X_j}||x_{im}-x_{jn}||.$$
$$h_{meanmax}^5(X_i,X_j)= \mathop{mean}_{x_{im}\in X_i} \mathop{max}_{x_{jn}\in X_j}||x_{im}-x_{jn}||.$$
$$h_{maxmax}^6(X_i,X_j)= \mathop{max}_{x_{im}\in X_i} \mathop{max}_{x_{jn}\in X_j}||x_{im}-x_{jn}||.$$

Apart from extending the Hausdorff distance by adding more operators, many incorrect measured cases can be avoided by taking $k$ nearest or farthest neighbors' average distance rather than just adopting one extreme case.
To make it a practice, we firstly define two functions $min\__k$ and $max\__k$ which are computing the $k_{th}$ largest and the $k_{th}$
smallest distance respectively.
To measure the distance between bag $X_i$ and $X_j$, we can implement this addition to modify the function as follows.

$$\bar{h}_{minmin}^1(X_i,X_j)= \mathop{min}_{x_{im}\in X_i} \frac{1}{k}\sum_{n=1}^{k} min\__k ||x_{im}-x_{jn}||.$$
$$\bar{h}_{meanmin}^2(X_i,X_j)= \mathop{mean}_{x_{im}\in X_i} \frac{1}{k}\sum_{n=1}^{k} min\__k ||x_{im}-x_{jn}||.$$
$$\bar{h}_{maxmin}^3(X_i,X_j)= \mathop{max}_{x_{im}\in X_i} \frac{1}{k}\sum_{n=1}^{k} min\__k ||x_{im}-x_{jn}||.$$
$$\bar{h}_{minmax}^4(X_i,X_j)= \mathop{min}_{x_{im}\in X_i}  \frac{1}{k}\sum_{n=1}^{k} max\__k ||x_{im}-x_{jn}||.$$
$$\bar{h}_{meanmax}^5(X_i,X_j)= \mathop{mean}_{x_{im}\in X_i}\frac{1}{k}\sum_{n=1}^{k} max\__k ||x_{im}-x_{jn}||.$$
$$\bar{h}_{maxmax}^6(X_i,X_j)= \mathop{max}_{x_{im}\in X_i}  \frac{1}{k}\sum_{n=1}^{k} max\__k ||x_{im}-x_{jn}||.$$

\begin{table*}[t]
\renewcommand\arraystretch{1.1}
    \centering
    \caption{The results on five benchmark data sets. The highest accuracies are highlighted in bold. }

    \label{tab:table1}
    \begin{tabular}{ccccccc}
      \toprule
      Algorithm & Musk1 & Musk2 & Elephant & Fox& Tiger & Average \\
      \midrule
      miBRV & 0.895 $\pm$ 0.078 & \textbf{0.930 $\pm$ 0.088}& \textbf{0.877 $\pm$ 0.102} & \textbf{0.670 $\pm$ 0.075}& \textbf{0.877 $\pm$ 0.102} & \textbf{0.851} \\

      miFV\cite{wei2014scalable} & \textbf{0.909 $\pm$ 0.089} & 0.884 $\pm$ 0.094 & 0.852 $\pm$ 0.081 & 0.621 $\pm$ 0.109& 0.813 $\pm$ 0.083 & 0.816 \\
      miGraph\cite{zhou2009multi} & 0.889 $\pm$ 0.073& 0.903 $\pm$ 0.086 & 0.869 $\pm$ 0.078 & 0.616 $\pm$ 0.079 & 0.801 $\pm$ 0.083 & 0.816\\
      MIBoosting \cite{MIBoosting2004} & 0.837 $\pm$ 0.120& 0.790 $\pm$ 0.088 & 0.827 $\pm$ 0.073 & 0.638 $\pm$ 0.102 & 0.784 $\pm$ 0.089 & 0.775\\
      miSVM \cite{andrews2002support}& 0.874 $\pm$ 0.120& 0.836 $\pm$ 0.088 & 0.822 $\pm$ 0.073 & 0.582 $\pm$ 0.102 & 0.789 $\pm$ 0.089 & 0.781\\
      EE-DD \cite{Zhang01em-dd:an}& 0.849 $\pm$ 0.098& 0.869 $\pm$ 0.108 & 0.771 $\pm$ 0.098 & 0.609 $\pm$ 0.101 & 0.730 $\pm$ 0.096 & 0.766\\
      MIWrapper \cite{2003MIWrapper}& 0.849 $\pm$ 0.106 & 0.796 $\pm$ 0.106 & 0.827 $\pm$ 0.088 & 0.582 $\pm$ 0.102 & 0.770 $\pm$ 0.092 & 0.765\\
      \bottomrule
    \end{tabular}

\end{table*}
The final representation of bag $X_i$, bag reference vector, is computed by a combination of these six distance operators, and denoted as

$$b^{X_i} = \left[ \bar{h}^t(X_i, X_j ) \right], \forall j \in [1,N], t \in [1,6].$$

Combining all or some of these six distance operators will extract more distinctive features so that we can gain  more comprehensive information for each bag with less training errors and improve the accuracy of the classifier.

\subsection{Bag Classification using Linear SVM}

As we can get a vector representation for a bag, we can use many existing classifiers for bag classification, such as SVM, Boosting, Random Forest.
For efficiency, we use SVM with linear kernel for bag training and bag label prediction.
The whole pipeline is illustrated in Algorithm~\ref{alg:1}. It consists with two steps, training and testing.
For both training and testing, we use the LibLinear \cite{fan2008liblinear} toolbox.

\begin{algorithm}[h]
        \caption{miBRV for bag training and classification.} \label{alg:1}
        \begin{algorithmic}[1] %line number
        \Require Data set \{$(X_1,Y_1),...,(X_N,Y_N)$\}

         \hspace{-1.3cm} \textbf{TRAIN:}
         \For{$i=1 \to N_{train}$}
         \State Map the original feature to Bag Reference Vector $b^{X_i}$ %= [h(X_i, X_1 ),...,h(X_i, X_j ),...,h(X_i, X_N)]$
         \State $b^{X_i}\gets b^{X_i}/||b^{X_i}||_2$
          \EndFor
          \State Use a linear SVM to train the transformed feature vectors \{$(b^{X_1},Y_1),...,(b^{X_i},Y_i),...,(b^{X_N},Y_N)$\} to learn a bag classifier B.

          \hspace{-1.3cm} \textbf{TEST:}

           \For {$j=1 \to N_{test}$}
           \State Map the original feature to Bag-Reference Vector $b^{X_j}$
           \State $b^{X_j}\gets b^{X_j}/||b^{X_j}||_2$
           \EndFor
        \Ensure The prediction of bag-level label B($b^{X_j}$).
        \end{algorithmic}
    \end{algorithm}

\section{Experiments} \label{section:Experiments}

\subsection{Benchmark Data sets}

In order to evaluate our method, we perform experiments on five benchmark data sets universally designed for MIL, including two Musk data sets ~\cite{dietterich1997solving} about molecule activity and three categories (elephant, fox, tiger) image data sets \cite{andrews2002support}.
In details, there are 47 positive and 45 negative bags in Musk1 while Musk2 are composed of 39 positive and 63 negative bags which are described by conformations with 166-dimensional feature vector.
On other three benchmark image data sets, each one is composed of 100 positive bags and 100 negative bags.
We perform training and testing for ten times by 10-fold cross-validation, and average classification accuracy and standard deviation of each class are reported.

Several popular MIL algorithms including the state-of-the-arts: miFV \cite{wei2014scalable}, miGraph \cite{zhou2009multi}, MIBoosting, miSVM \cite{andrews2002support}, EM-DD \cite{Zhang01em-dd:an}, and MIWrapper \cite{2003MIWrapper}, are referred  for comparison to evaluate our results.
As shown in Table~\ref{tab:table1}, it indicates that miBRV are so competitive that it achieves the highest performance except on the MUSK1 data set.
% Hence, the average accuracy for the five data sets has been improved a lot by miBRV.
The average accuracy of miBRV over the five data sets has been improved by 3.5\% to a large margin when comparing to the latest miFV method.
%The quantitative performance figures of the compared methods are quoted from the previous literatures.
The excellent results clearly demonstrate that miBRV is robust and can extract the most effective representation for a bag in MIL problems.

\subsection{Text Categorization}

Besides the benchmark tasks, the text categorization is another common application of MIL. For better comparison, we take the same twenty data sets derived from the 20 Newsproups corpus as in \cite{zhou2009multi}. In each category, there are 100 bags among which half bags are positive and others are negative. In addition, each instance is a post represented by the top 200 TF-IDF features.

In the same way, we carry out experiments on this data set using 10-fold cross-validation and report the average accuracy in Table~\ref{tab:table2}.
On this occasion, comparisons have been made between our miBRV, MI-Kernel and miGraph on these text categorization tasks.
On 13 data sets out of 20, miBRV achieves the superior performance. The best average accuracy over all data sets indicates that the miBRV outperforms other two competing algorithms, miGraph and MI-Kernel \cite{MIkernel2002}.

\begin{table}[h]
%\footnotesize
    \caption{The results of twenty data sets of text categorization.}
    \label{tab:table2}
    \begin{tabular}{lccc}
      \toprule
Data set&MIkernel&miGraph&miBRV\\
\midrule
alt.atheism& 60.2  &65.5  & \textbf{77.0} \\
comp.graphics& 47.0  & \textbf{77.8 } & 72.1\\
\small{comp.os.ms-windows.misc}& 51.0 & 63.1  &\textbf{64.1}\\
\small{comp.sys.ibm.pc.hardware}&  46.9  & 59.5  & \textbf{69.0} \\
comp.sys.mac.harware&  44.5  & 61.7  & \textbf{70.7}  \\
comp.window.x&  50.8  & 69.8  & \textbf{80.7} \\
misc.forsale&  51.8 & 55.2  & \textbf{61.2}\\
rec.autos &  52.9  & \textbf{72.0} & 64.1\\
rec.motorcycles &  50.6 & \textbf{64.0 } & 54.4\\
rec.sport.baseball &  51.7  & 64.7  & \textbf{77.8}\\
rec.sport.hockey &  51.3  & 85.0  & \textbf{85.0}\\
sci.crypt &  56.3  & 69.6  & \textbf{70.3}\\
sci.electronics &  50.6  & 87.1 & \textbf{90.7}\\
sci.med &  50.6 & 62.1  & \textbf{74.8}\\
sci.space &  54.7 & \textbf{75.7 } & 67.8 \\
sci.religion.christian &  49.2  & 59.0 & \textbf{68.6}\\
talk.politics.guns &  47.7  & 58.5 & \textbf{66.2}\\
talk.politics.mideast &  55.9  & \textbf{73.6} & 65.1\\
talk.politics.misc &  51.5 & \textbf{70.4} & 63.8\\
talk.religion.misc &  55.4 & \textbf{63.3} & 60.8\\
\midrule
Average &51.5 & 67.8 & \textbf{70.1} \\
  \bottomrule
   \end{tabular}
\end{table}

\begin{table*}[t]
\renewcommand\arraystretch{1.1}
    \centering
    \caption{The results on five benchmark data sets of parameter analysis. The distance operator is a combination of six operators.}
    \label{tab:table3}
    \begin{tabular}{ccccccc}
      \toprule
       Parameters& Musk1 & Musk2 & Elephant & Fox& Tiger  \\
      \midrule
     $k$=1 & \textbf{0.882 $\pm$ 0.088 }&\textbf{0.907 $\pm$ 0.095} & 0.849 $\pm$ 0.076 & 0.623 $\pm$ 0.098 & 0.829 $\pm$ 0.076  \\

     $k$=2 & 0.870 $\pm$ 0.100 & 0.901 $\pm$ 0.098 & \textbf{0.850 $\pm$ 0.067} & 0.669 $\pm$ 0.101& \textbf{0.841 $\pm$ 0.080}
       \\
     $k$=3 & 0.862 $\pm$ 0.075 & 0.893 $\pm$ 0.102 & 0.838 $\pm$ 0.075 & \textbf{0.670 $\pm$ 0.098}& 0.815 $\pm$ 0.085  \\
     $k$=4 & 0.860 $\pm$ 0.111 & 0.895 $\pm$ 0.093 & 0.831 $\pm$ 0.070 & 0.662 $\pm$ 0.091& 0.816 $\pm$ 0.083 \\
      \bottomrule
    \end{tabular}
\end{table*}

\begin{table*}[t]
\renewcommand\arraystretch{1.2}
    \centering
    \caption{The results on five benchmark data sets of parameter analysis.The value of $k$ is 2.}
    \label{tab:table4}
    \begin{tabular}{ccccccc}
      \toprule
       Parameters& Musk1 & Musk2 & Elephant & Fox& Tiger  \\
      \midrule

       %K=1,$[b_{h^2}^X,b_{h^5}^X]$&
%       0.886 $\pm$ 0.089 & 0.930 $\pm$ 0.088& 0.843 $\pm$ 0.070 & 0.611 $\pm$ 0.112& 0.847 $\pm$ 0.069 & 0.823 \\
%
%       K=1,$[b_{h^1}^X,b_{h^3}^X]$&
%        0.895 $\pm$ 0.078 & 0.917 $\pm$ 0.082& 0.794 $\pm$ 0.085 & 0.670 $\pm$ 0.102& 0.793 $\pm$ 0.092 & 0.805 \\

     $[b_{\bar{h}^1}^X,b_{\bar h^3}^X]$&
     0.872 $\pm$ 0.094 & 0.909 $\pm$ 0.083& 0.797 $\pm$ 0.075 & \textbf{0.670 $\pm$ 0.111}& 0.803 $\pm$ 0.085  \\

     $[b_{\bar{h}^2}^X,b_{\bar h^5}^X]$&
     \textbf{0.886 $\pm$ 0.092} & \textbf{0.930 $\pm$ 0.088} & 0.843 $\pm$ 0.076 & 0.611 $\pm$ 0.113& 0.847 $\pm$ 0.085  \\
     %K=1,$[b_{h^2}^X,b_{h^4}^X,b_{h^5}^X]$ & 0.890 $\pm$ 0.092 & 0.921 $\pm$ 0.098 & 0.864 $\pm$ 0.070 & 0.609 $\pm$ 0.113& 0.858 $\pm$ 0.072 & 0.828 \\

     $[b_{\bar h^2}^X,b_{\bar h^4}^X,b_{\bar h^5}^X]$ & 0.880 $\pm$ 0.102 & 0.903 $\pm$ 0.100 & \textbf{0.877 $\pm$ 0.071} & 0.640 $\pm$ 0.102& \textbf{0.877 $\pm$ 0.067}  \\

     %K=1,$[b_{h^1}^X,...,b_{h^6}^X]$ & 0.882 $\pm$ 0.088 & 0.907 $\pm$ 0.095 & 0.849 $\pm$ 0.076 & 0.623 $\pm$ 0.098 & 0.829 $\pm$ 0.076 & 0.818 \\

      $[b_{\bar h^1}^X,...,b_{\bar h^6}^X]$ & 0.870 $\pm$ 0.100 & 0.901 $\pm$ 0.098 & 0.850 $\pm$ 0.067 & 0.669 $\pm$ 0.101& 0.841 $\pm$ 0.080  \\

      \bottomrule
    \end{tabular}
\end{table*}

\subsection{Parameters Discussion}

To deeply investigate miBRV, we discuss two main parameters in miBRV in this subsection.
As illustrated in Section~\ref{section:Method}, we generate the final vector by combining different affinity matrices which are mapped by different distance operators together such as $[b_{\bar{h}^1}^X,...,{b}_{\bar{h}^6}^X]$ or just selecting some of them.
In addition, the value of $k$, the number of averaging neighbors to be adopted, is a significant parameter for our experiment as well.

At first, we keep one factor, distance operator, unchanged  with $k$ changing form 1 to 4.
The part of results for $[b_{\bar{h}^1}^X,b_{\bar{h}^2}^X,b_{\bar{h}^3}^X,b_{\bar{h}^4}^X,b_{\bar{h}^5}^X,{b}_{\bar{h}^6}^X]$ are shown in Table \ref{tab:table3}.
These results illustrate that increasing $k$ ameliorates the accuracy on Elephant and Fox but experiences a decline on Musk data sets at the same time. As a whole, it reaches the acme of average accuracy at $k$=2.

Then we fix $k$ to 2 and test some different combinations of distance functions to extract a more informative feature vectors for diverse cases.
Generally, higher dimensions feature vector improves the performance of classifier. And the more distance operators we used, the more robust miBRV is as a feature vector.
Table~\ref{tab:table4} contains some details of the results with different parameters, from which we can find that, although results changes a lot from different parameters, most averaging accuracy are competitive with the state-of-the-art algorithms.
The best performance of each column is bolded.

\section{Conclusions} \label{section:Con}

In this paper, we propose a novel technique for Multi-instance Learning. We focus on the inherent information on each bag, trying to delineate it by computing the similarity to other bags.
In addition, our diverse distance definition fits it well, considering the crux of MIL that the proportion of positive instances in  positive bags is ambiguous.
No previous works adopt this straightforward but efficacious feature representation method. And the performances of our algorithm on these data sets popularly used for emulating MIL algorithms are superior to the state-of-the-art algorithms. What's more, the proposed method produces a very simple vector representation for a bag, which works well with a linear SVM. Both the methodology and experimental results of the proposed approach show that it is very robust and effective.

In the future, on one hand, we may extend our method by changing the choice of reference bags. For instance, we can generate a great deal of reference bags in which the instances are randomly selected from the original bags. By this way we may describe our bags more accurately with more references if we can solve its possible computational expense. On the other hand, hewing to the intrinsic characteristic of MIL, we can extract features which describe the relationship of the instances in each bag by means of, for instance, adding some mathematical statistics such as standard deviation of instances in a bag, allowing us to distinguish different bags more clearly to solve the core problem of MIL tasks.

\section*{Acknowledgments}
This work was primarily supported by National Natural Science Foundation of China (NSFC) (No. 61503145).

{\small
\bibliographystyle{ieee}
\bibliography{egbib}

\begin{thebibliography}{10}\itemsep=-1pt

\bibitem{andrews2002support}
S.~Andrews, I.~Tsochantaridis, and T.~Hofmann.
\newblock Support vector machines for multiple-instance learning.
\newblock In {\em NIPS}, 2002.

\bibitem{Babenko}
B.~Babenko, N.~Verma, P.~Dollar, and S.~Belongie.
\newblock Multiple instance learning with manifold bag.
\newblock In {\em ICML}, 2011.

\bibitem{babenko2011robust}
B.~Babenko, M.-H. Yang, and S.~Belongie.
\newblock Robust object tracking with online multiple instance learning.
\newblock {\em IEEE Transactions on Pattern Analysis and Machine Intelligence},
  33(8):1619--1632, 2011.

\bibitem{cinbis2014multi}
R.~G. Cinbis, J.~Verbeek, and C.~Schmid.
\newblock Multi-fold mil training for weakly supervised object localization.
\newblock In {\em CVPR}, 2014.

\bibitem{Deselaers_aconditional}
T.~Deselaers and V.~Ferrari.
\newblock A conditional random field for multiple-instance learning.
\newblock In {\em ICML}, 2010.

\bibitem{dietterich1997solving}
T.~G. Dietterich, R.~H. Lathrop, and T.~Lozano-P{\'e}rez.
\newblock Solving the multiple instance problem with axis-parallel rectangles.
\newblock {\em Artificial intelligence}, 89(1):31--71, 1997.

\bibitem{2003MIWrapper}
E.T.Frank and X.Xu.
\newblock Applying propositional learning algorithms to multi-instance data.
\newblock {\em University of Waikato,Department of Comuter Science,University
  of Waikato, Hamilton, NZ, Tech. Rep.}, 2003.

\bibitem{fan2008liblinear}
R.-E. Fan, K.-W. Chang, C.-J. Hsieh, X.-R. Wang, and C.-J. Lin.
\newblock Liblinear: A library for large linear classification.
\newblock {\em The Journal of Machine Learning Research}, 9:1871--1874, 2008.

\bibitem{felzenszwalb2010object}
P.~F. Felzenszwalb, R.~B. Girshick, D.~McAllester, and D.~Ramanan.
\newblock Object detection with discriminatively trained part-based models.
\newblock {\em IEEE Transactions on Pattern Analysis and Machine Intelligence},
  32(9):1627--1645, 2010.

\bibitem{MIkernel2002}
T.~G{\"a}rtner, P.~A. Flach, A.~Kowalczyk, and A.~J. Smola.
\newblock Multi-instance kernels.
\newblock In {\em ICML}, 2002.

\bibitem{huttenlocher1993comparing}
D.~P. Huttenlocher, G.~Klanderman, W.~J. Rucklidge, et~al.
\newblock Comparing images using the hausdorff distance.
\newblock {\em IEEE Transactions on Pattern Analysis and Machine Intelligence},
  15(9):850--863, 1993.

\bibitem{liu2012key}
G.~Liu, J.~Wu, and Z.-H. Zhou.
\newblock Key instance detection in multi-instance learning.
\newblock 2012.

\bibitem{maron1998multiple}
O.~Maron and A.~L. Ratan.
\newblock Multiple-instance learning for natural scene classification.
\newblock In {\em ICML}, 1998.

\bibitem{parikh2011relative}
D.~Parikh and K.~Grauman.
\newblock Relative attributes.
\newblock In {\em ICCV}, 2011.

\bibitem{shen2013face}
W.~Shen, B.~Wang, Y.~Wang, X.~Bai, and L.~J. Latecki.
\newblock Face identification using reference-based features with message
  passing model.
\newblock {\em Neurocomputing}, 99:339--346, 2013.

\bibitem{Viola06multipleinstance}
P.~Viola, J.~C. Platt, and C.~Zhang.
\newblock Multiple instance boosting for object detection.
\newblock In {\em NIPS}, 2006.

\bibitem{wang2013one}
X.~Wang, Z.~Zhang, Y.~Ma, X.~Bai, W.~Liu, and Z.~Tu.
\newblock One-class multiple instance learning via robust pca for common object
  discovery.
\newblock In {\em ACCV}. 2013.

\bibitem{wang2014robust}
X.~Wang, Z.~Zhang, Y.~Ma, X.~Bai, W.~Liu, and Z.~Tu.
\newblock Robust subspace discovery via relaxed rank minimization.
\newblock {\em Neural computation}, 26(3):611--635, 2014.

\bibitem{wei2014scalable}
X.-S. Wei, J.~Wu, and Z.-H. Zhou.
\newblock Scalable multi-instance learning.
\newblock In {\em ICDM}, 2014.

\bibitem{MIBoosting2004}
X.~Xu and E.~Frank.
\newblock Logistic regression and boosting for labeled bags of instances.
\newblock In {\em Knowledge Discovery and Data Mining}, pages 272¨C--281. Proc.
  8th Pacific-Asia Conf, 2004.

\bibitem{zha2008joint}
Z.-J. Zha, X.-S. Hua, T.~Mei, J.~Wang, G.-J. Qi, and Z.~Wang.
\newblock Joint multi-label multi-instance learning for image classification.
\newblock In {\em CVPR}, 2008.

\bibitem{zhang2013real}
K.~Zhang and H.~Song.
\newblock Real-time visual tracking via online weighted multiple instance
  learning.
\newblock {\em Pattern Recognition}, 46(1):397--411, 2013.

\bibitem{Zhang01em-dd:an}
Q.~Zhang and S.~A. Goldman.
\newblock Em-dd: An improved multiple-instance learning technique.
\newblock In {\em NIPS}, 2001.

\bibitem{zhou2009multi}
Z.-H. Zhou, Y.-Y. Sun, and Y.-F. Li.
\newblock Multi-instance learning by treating instances as non-iid samples.
\newblock In {\em ICML}, 2009.

\end{thebibliography}
}

\end{document}